\documentclass[conference]{IEEEtran}
\IEEEoverridecommandlockouts
\usepackage{cite}
\usepackage{savesym}
\usepackage{amsmath,amssymb,amsfonts}
\savesymbol{iint}
\usepackage{txfonts}
\restoresymbol{TXF}{iint}
\usepackage{algorithmic}
\usepackage{graphicx}
\usepackage{textcomp}
\usepackage{xcolor}
\usepackage{textgreek}
\usepackage{mathtools}
\usepackage{wrapfig}
\usepackage{hyperref}
\usepackage{fancyhdr}

\fancypagestyle{plain}{%
   \fancyhf{} 
   \fancyfoot[C]{\small {Source Code: {https://tinyurl.com/v7bbrkb} }}}

\makeatletter
\newcommand{\newlineauthors}{%
  \end{@IEEEauthorhalign}\hfill\mbox{}\par
  \mbox{}\hfill\begin{@IEEEauthorhalign}
}

\def\BibTeX{{\rm B\kern-.05em{\sc i\kern-.025em b}\kern-.08em
    T\kern-.1667em\lower.7ex\hbox{E}\kern-.125emX}}
\begin{document}

\title{Textureless Object Recognition: An Edge-based Approach}

\author{\IEEEauthorblockN{1\textsuperscript{st} Frincy Clement}
\IEEEauthorblockA{\textit{dept. of Computing Science} \\
\textit{University of Alberta}\\
Edmonton, Canada \\
frincy@ualberta.ca}
\and
\IEEEauthorblockN{2\textsuperscript{nd} Kirtan Shah}
\IEEEauthorblockA{\textit{dept. of Computing Science} \\
\textit{University of Alberta}\\
Edmonton, Canada \\
knshah@ualberta.ca}
\and
\IEEEauthorblockN{3\textsuperscript{rd} Dhara Pancholi}
\IEEEauthorblockA{\textit{dept. of Computing Science} \\
\textit{University of Alberta}\\
Edmonton, Canada \\
dhara@ualberta.ca}

\newlineauthors
\IEEEauthorblockN{4\textsuperscript{th} Gabriel Lugo Bustillo}
\IEEEauthorblockA{\textit{dept. of Computing Science} \\
\textit{University of Alberta}\\
Edmonton, Canada \\
lugobust@ualberta.ca}
\and
\IEEEauthorblockN{5\textsuperscript{th} Dr. Irene Cheng}
\IEEEauthorblockA{\textit{dept. of Computing Science} \\
\textit{University of Alberta}\\
Edmonton, Canada \\
locheng@ualberta.ca}
}

\maketitle

\begin{abstract}
Textureless object recognition has become a significant task in Computer Vision with the advent of Robotics and its applications in manufacturing sector. It has been challenging to obtain good accuracy in real time because of its lack of discriminative features and reflectance properties which makes the techniques for textured object recognition insufficient for textureless objects. A lot of work has been done in the last 20 years, especially in the recent 5 years after the TLess and other textureless dataset were introduced. In this project, by applying image processing techniques we created a robust augmented dataset from initial imbalanced smaller dataset. We extracted edge features, feature combinations and RGB images enhanced with feature/feature combinations to create 15 datasets, each with a size of ~340,000. We then trained four classifiers on these 15 datasets to arrive at a conclusion as to which dataset performs the best overall and whether edge features are important for textureless objects. Based on our experiments and analysis, RGB images enhanced with combination of 3 edge features performed the best compared to all others. Model performance on dataset with HED edges performed comparatively better than other edge detectors like Canny or Prewitt. 
\end{abstract}


\section{Introduction}
In recent years, a lot of labor-intensive activities in manufacturing sector has been automated using robots. In industrial assembly line, a robotic arm is used to pick up small objects and place it in its location. These industrial objects are mostly textureless and are different from textured objects in terms of lack of discriminative features. Often, they are smaller in size and can be similar to each other. In real world, they are present along with multiple objects of different classes in different scene configurations, which makes the recognition even more challenging. 

In the early 2000s, detection and recognition of the textureless objects seemed to be a daunting issue, even for the state-of-art detectors at that time, like SIFT (Scale Invariant Feature Transform) and SURF (Speeded UP Robust Features). This was due to their dependence on the high-featured, texture-rich informative regions which were sparse in the textureless occurrences. Ultimately, this led to several works which contributed for the detection and recognition of textureless objects, which can be thus classified into three categories: View-based, Feature-based, and Shape-based.

\subsection{View-based} 

View-based methods work by comparing the object of interest (OOI) and the pre-calculated 2D views of that object, denoted by aspect graphs. This method can provide a rough estimate of the pose of the object, but it suffers from computational load of searching a large spatial domain. To overcome this issue, Cyr and Kimia ~\cite{cyr2004similarity} proposed a method that grouped similar OOI together under a single class and similarity scores were calculated to perform recognition. Eggert et al. \cite{eggert1993scale}, Ulrich et al. \cite{ulrich2011combining} and Steger et al. \cite{steger2002occlusion} introduced various improvements on the aspect graphs and similarity metric. This approach, however was not widely used because of its high complexity. 

\subsection{Feature based method}

Tombari et al. \cite{tombari2013bold} devised a method called BOLD in which they attempt to tackle the issue of object detection under clutter. In this method, neighbouring line segments were aggregated to form a feature representation of the object. These line segments are invariant to rotation, translation as well as scaling. A limitation of BOLD is that it fails if there is clutter in the line segment selected for feature description.

Intended for object recognition and tracking using an RGB-D sensor, Jiang et al. \cite{jiang2016object} proposed a method that detects and recognises small objects. The image is divided into various parts using image segmentation. Then it is sent to a classifier as an input which classifies the object using the features extracted in the earlier stages. 

Hodan et al. \cite{hodavn2015detection} proposed a technique for detecting textureless objects using AR, which used an edge-based detector which gave better results than canny edge detector. And efficient towards the object detection as it searched edgelet in stripes instead of wedges. 

While the above-mentioned methods do a commendable job of detecting textureless objects, they were computationally expensive. In 2016 Jacob et al. \cite{chan2016border} introduced BORDER which is an acronym for Bounded Oriented Rectangle Descriptors for Enclosed Region, in which they use a rectangle to encapsulate the object so that minimum outliers are detected. It outperformed the previous state-of-the-art descriptors BOLD and Line2D when compared for images having occlusion and clutter. 

Counting the number of textureless objects present in a scene was still a challenge until Verma et al. \cite{verma2016vision} proposed a method which counts the objects using shape and colour feature which are obtained through morphological boundary extraction and segmentation. The uniqueness in this method was it was invariant to scaling and rotation. Further, different models such as SVM, KNN etc were used for the classification. 

To improve accuracy, Thoduka et al. \cite{thoduka2016rgb} in 2016 came up with an approach which used RGB-D based features to recognize objects in the industrial assembly line such as nuts, bolts, bearing large and small black profiles etc.The approach performed efficiently on most of the objects in dataset but they had poor performance recognizing smaller objects.

Following upon his scholarly work, Hodan et al. \cite{hodan2017t} improved upon his previous research by presenting a new dataset in 2017, “T-less” publicly for better understanding of 3D object from different dimensions. This dataset contains industrial objects which have no texture while having feature similarities of shape/size which makes it quite hard to differentiate. 

To overcome the short comings of previous methods, BIND \cite{chan2017bind} was introduced during 2017. It is a detector that uses binary nets having multiple layer for accurate detection of textureless object. To improve the speed of object detection, in 2018 Hancheng et al. \cite{yu2018fast} introduced a faster approach to detect textureless objects even in the cluttered background and transformations. 

Fang et al. \cite{fang2019dog} deduced that for object recognition, the background of an image affects the performance of neural networks, and hence came up with a method called DOG, built upon CNN which effectively removed the background of an input image for improving the efficiency of the classifier. 

\subsection{Shape Based Approach}

Template matching is a popular method in textureless object recognition, dating back as early as the 2000s in lieu of the fact that it provides high mAP and close to real-time detection. But, up until now, templates were generated using one modality only (depth or colour or point cloud or 2D). However, in 2011 Hinterstoisser et al. \cite{hinterstoisser2011multimodal} put forth the idea of multi-modal template matching. The template comprises of both the image cue (colour information) as well as depth cue (3D) information. The method can detect objects in real-time as well as under heavy clutter.A caveat of the above approach is that it is not invariant to translation, rotation, or scaling.

In the following year, Hinterstoisser et al. \cite{hinterstoisser2011gradient} overcame the afore-mentioned issue by considering all the gradient orientations for a template instead of just the dominant ones. Due to this, the method was more robust to translations. While it does suppress the caveat of the above approach to a certain extent, it is still not invariant to scaling. 

In 2015 Hodan et al. \cite{hodavn2015detection} came up with a very efficient way of template matching, resolving a lot of its computational complexity in performing searches of each sliding window with each stored template. This method resulted in real time performance with low false rates and multiple object detection.

In 2018, an edge-based Hierarchical Template Matching algorithm was proposed by Tsai et al. \cite{tsai2018real} based upon Line2D algorithm, which gave an efficient method for real-time object detection and recognition. As per this paper, edges are identified as the most stable feature for a textureless object. This was the baseline information for building our project.

2018 saw many approaches for textureless object detection. PoseCNN (current state-of-the-art), an approach by Xiang et al., \cite{xiang2017posecnn} is a convolutional network for predicting object pose with varied orientations. It predicts object’s translation matrix by means of calculating object’s distance with respect to the camera. They also calculate the object’s rotation matrix. Midway through 2019, Park et al. \cite{park2019multi} proposed a method that takes depth image as an input and a region of interest (ROI) is generated. 

In the next section, we introduce the existing work done in the field of textureless object recognition by using a combination of image processing and deep learning techniques. Our proposed method is explained step by step in Section III. Furthermore, Section IV details the experiments performed, results and analysis. The report ends with Conclusion and future work, Acknowledgement, Author Biography and References.

\section{Related Work}

Our proposed work uses image processing techniques combined with machine learning techniques for textureless object recognition. We perform image processing techniques to extract features and create multiple training sets. We compare these datasets using machine learning techniques. In this section, previous researches which uses similar approach are summarized. 

Jiang et al. \cite{jiang2016object} proposed an RGB-D based method for the detection of various types of common objects. They used depth filtering alongside clustering, coupled with the watershed algorithm on the depth data for the target object recognition. Once recognized, they apply a tracking method to reduce the search space and the computational load. Further, the method was evaluated using a random forest classifier. However, for real-time object detection, only RGB dataset was used. 

Wang et al. \cite{wang2016self} proposed a method for object recognition using model-based CNN learning. They introduced a unified structure composed of automatic object reconstruction and multitask learning aimed at doing object recognition by model-based training alone. This method reduced the gap between real images and synthetically rendered images for better accuracy. 

Later, a technique to count textureless objects with shape and colour feature was developed by Verma et al. \cite{verma2016vision}.  First, the shape and color features were extracted during pre-processing using morphological boundary extraction and segmentation via mean hue value (colour features) and Hu-moments (shape features). Next, SVM, kNN, NN and tree bagging were applied for the classification of objects. Through experiments, they observed that tree bagging has the best accuracy of all the classifiers. Finally, the classified objects were counted by drawing bounding boxes around them.

Thoduka et al. \cite{thoduka2016rgb} put forth a method for object detection using 3D point cloud.  They defined size, shape, and color of objects by using various metrics including the bounding box. The extracted features were then passed through SVM for classification. However, the accuracy of the method is inversely proportional to the size of the object in each image.

A breakthrough in the field of textureless object recognition happened when Hodan et al. \cite{hodan2017t} introduced T-less dataset for 6D pose estimation. The dataset houses thirty industry-relevant textureless objects with no discriminative colour or illumination properties. They are not symmetrical with respect to shape or size. It contains 39K training images along with 10K testing images that exhibit occlusion and clutter. It also has manually created 3D CAD model, as well as one which is constructed semi-automatically. For our project, we used the 2D images taken from 3D printed models of this dataset as our baseline data.

Hodan et al. \cite{hodavn2015detection} came up with a method using purely edge detector which uses edgelet strips in place of wedges. When the binary map of edges is provided, this method takes the edglets of equal fix lengths and they are represented by their midpoints and orientation, which helps in tracking the orientation. Two images are matched using their distance transformation in orientation which proves the presence of that object in image even if more than one same object is present.

A technique called PoseCNN introduced by Xiang et al. \cite{xiang2017posecnn} used for 6D pose estimation, estimated the 3D translation and rotation of an object by predicting the center of the object’s distance from the camera. For this process, they also introduced a novel loss function called P loss for objection detection with clutter. Speaking of which, clutter and background are important aspects when it comes to image segmentation. Fang et al. \cite{fang2019dog} devised a method called DOG which removes the background of the images to improve the object detection accuracy. The method uses DCGAN for classification.

Tsai et al. \cite{tsai2018real} came up with a new approach for Template matching which proved to be an efficient method for object detection and recognition. They chose edges as they were identified as the most stable feature for a textureless object. After image enhancement, the input image goes through the edge-based template matching algorithm. It detects the color edge by finding the one with maximum magnitude across all color channels. These detected edges are then quantized into one of 8 bins and one-hot encoded for computational efficiency. The one-hot encoded edge orientations are then spread using dilation.  The algorithm then checks for similarity with the precomputed database which has edge templates of reference images and lookup tables of edge orientations, which will help in efficient matching and calculation of similarity score. 

For the detection of textureless objects, the importance of depth images over colour images cannot be overstated. Park et al. \cite{park2019multi} developed a method called Multi-task template matching for object detection, segmentation and pose estimation using depth images. They put forth a method to match the Nearest Neighbour template using a similarity score. Since shared feature maps were used to predict the segmentation mask and pose of an object, additional training was not required for a new object. For real-time object detection and recognition, Region of Interest (ROI) predicted by ResNet-50 architecture pretrained offline on the templates, were compared with the ROI of test image predicted real-time by another ResNet-50.

\section{Proposed Method}

\begin{figure*}[htbp]
\centerline{\includegraphics[width=1\textwidth]{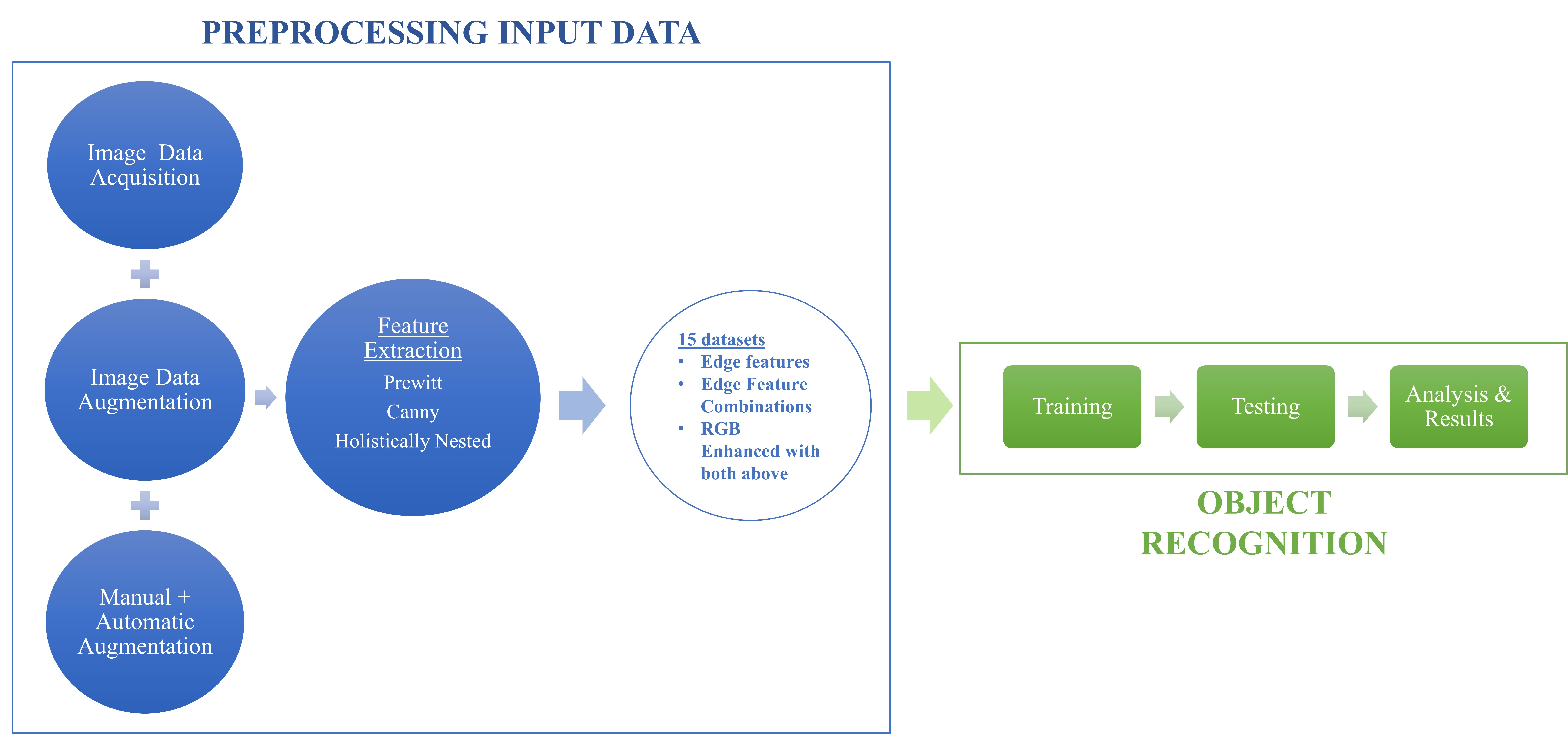}}
\caption{Project Workflow}
\label{fig:Fig1}
\end{figure*}

Our work focuses on recognition of textureless objects in 30 classes. In this section, we will discuss how we acquired the intial dataset, what image processing techniques are applied on the input images to convert it into features and feature-boosted versions of original images and which classifiers are selected for object recognition. The ``Fig.~\ref{fig:Fig1}'' depicts the block diagram of our proposed method. 

\medskip

\noindent \textbf{Project Implementation Structure} 

\medskip

Our project was implemented in 3 parts. In Part 1, we focused on applying image processing techniques to create a balanced augmented dataset from ground truth. In Part 2, we apply edge operators to create 14 new datasets to perform feature comparison. In Part 3, we train a classifier and predict results.

\noindent \textit{Part 1: Creation of Initial Augmented Data}	
\begin{itemize}
    \item Image Data Acquisition
    \item Obtaining Ground Truth data
    \item Data Balancing and Augmentation
    \begin{itemize}
        \item Manual: Image Processing Techniques
        \item Automatic: Augmentor API
    \end{itemize}
\end{itemize}

\noindent \textit{Part 2: Creation of 14 new datasets for feature comparison}
\begin{itemize}
    \item	Creation of Feature Only Dataset
    \item	Creation of Feature Combinations
    \item	Creation of Feature Enhanced RGB Dataset
    \item	Creation of Feature Combination Enhanced RGB Dataset
    
\end{itemize}        

\noindent \textit{Part 3: Training and Testing on Multiple Classifiers}

\medskip 

\vspace{35mm}

\noindent \textbf{Part 1: Creation of Initial Augmented Data}

\medskip

\subsubsection{Image Data Acquisition}

The dataset we used for our experiments are based on TLess Dataset \cite{hodan2017t} introduced by Hodan et al. 30 textureless objects were 3-D printed from the CAD model and using an Intel RealSense camera, 3D images were captured at the Multimedia Research Lab, University of Alberta. We used the 3D-rendered 2D images to start our implementation for object recognition. 

\subsubsection{Obtaining Ground Truth Data}

\begin{figure*}[htbp]
\centerline{\includegraphics[width=1\textwidth]{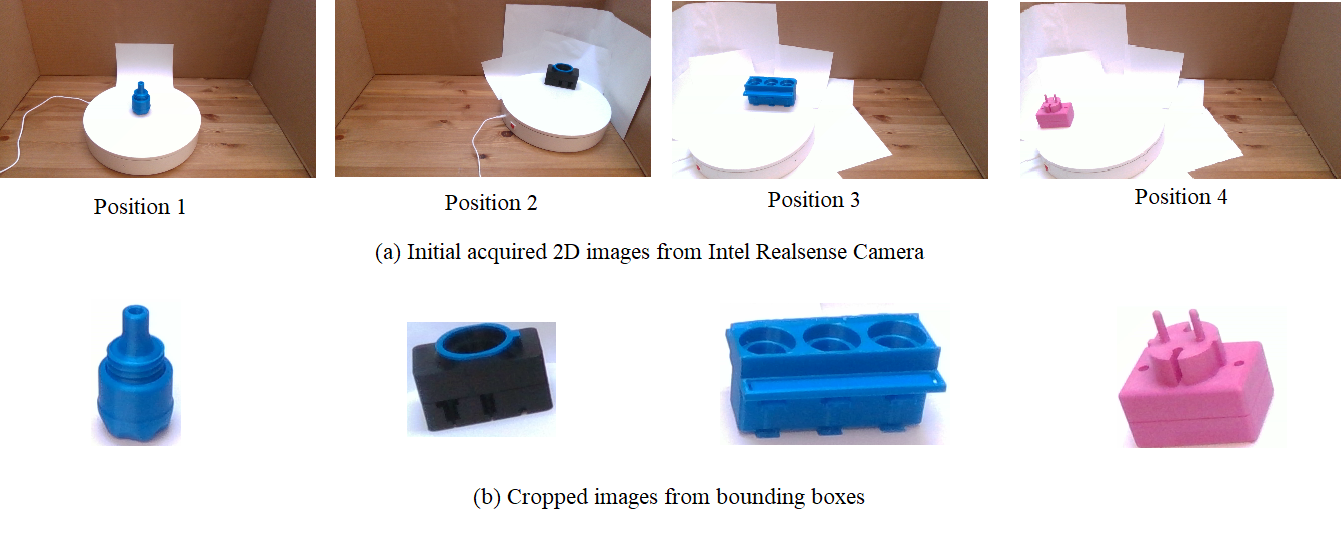}}
\caption{Data Preprocessing to obtain ground truth objects from initial acquired 2D images}
\label{fig:Fig2}
\end{figure*}

We started our experiments with ~27000 images that contained textureless objects from 30 different classes. The objects that were captured covered only small portion of the images. We used regionprops() from sci-kit learn library to obtain bounding boxes for each object. From those bounding boxes, the objects were cropped leaving few pixels on all sides. The bounding box information was also saved as annotations for training any of YOLO versions in the form: Classname, x-center, y-center, height, width – all normalized to image height and width. 
The different sample images captured from Intel RealSense camera and the output of regionprops() are shown in ``Fig.~\ref{fig:Fig2}''.

\subsubsection{Data Balancing \& Augmentation}

\begin{figure*}[htbp]
\centerline{\includegraphics[width=0.8\textwidth]{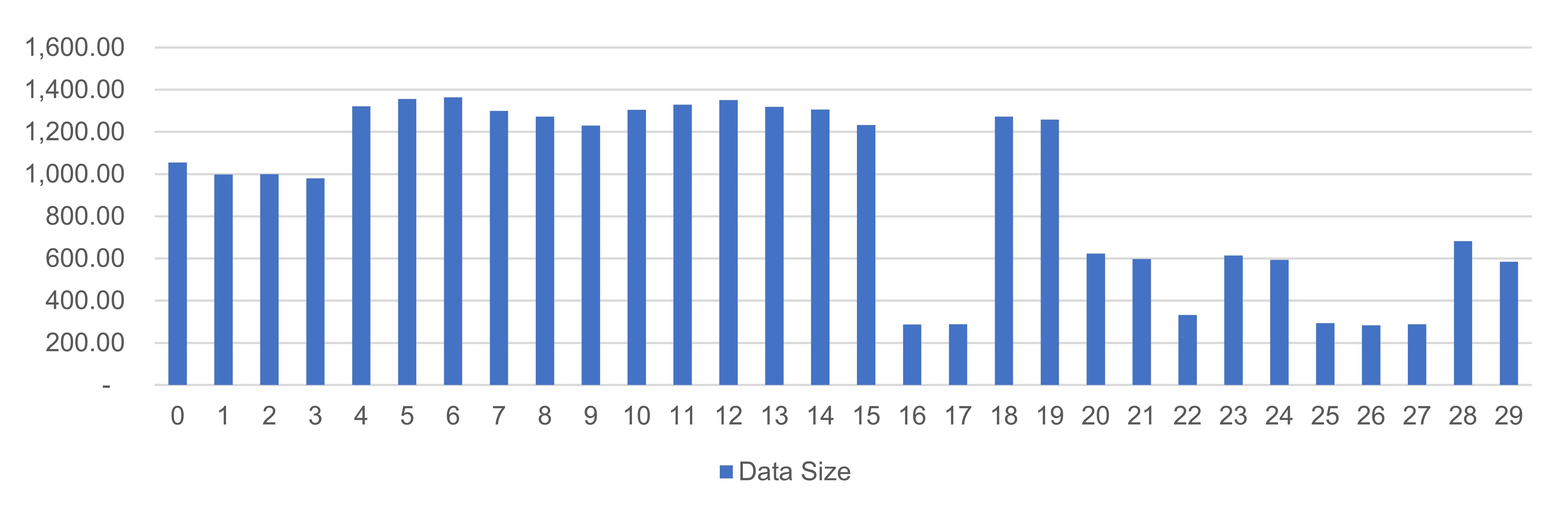}}
\caption{Distribution of data across 30 classes}
\label{fig:Fig3}
\end{figure*}

The data provided was captured in 4 different orientations for most of the classes except for the ones that had unique features. The data was unevenly distributed among 30 different classes. As seen on ``Fig.~\ref{fig:Fig3}'' Whereas most of the classes had \~1200 images, some had only \~300.

To balance the dataset, we calculated a balancing factor. While applying manual augmentation, we considered this balancing factor to remove the imbalance in dataset. 

\begin{figure*}[htbp]
\centerline{\includegraphics[width=0.7\textwidth]{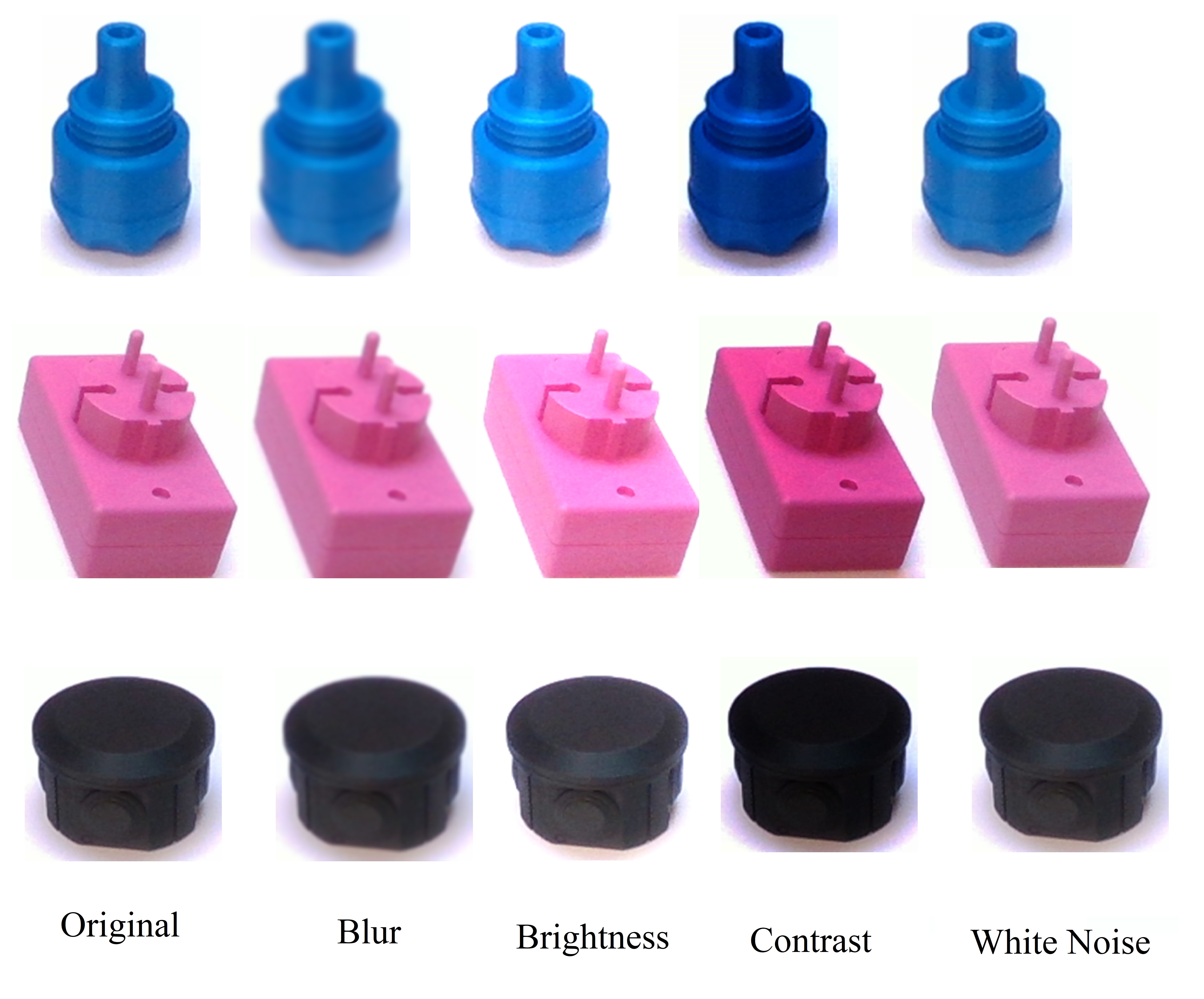}}
\caption{Results of Manual Augmentation}
\label{fig:Fig4}
\end{figure*}

The number of images captured using Intel RealSense is 1000 per object, which is too small to efficiently train a classifier. Hence, we need to apply image augmentation to generate artificial data. We augmented the input image in two ways: Manually applying image processing techniques and using Augmentor API.  

\noindent a)	Manual Augmentation: As proposed in the recognition method by Huang et al. \cite{huang2019rapid}, we applied the following techniques on the ground truth data with a balancing factor. The results are shown in ``Fig.~\ref{fig:Fig4}''

\begin{enumerate}
  \item Contrast Enhancement – to improve the contrast of the textureless object with respect to the background
  \item Addition of Noise – to avoid overfitting of input data
  \item Brightness transformation – to simulate the object detection in the lighting in manufacturing facility
  \item Addition of Blur – to account for the blurring that can be caused by the camera movement
\end{enumerate}

\noindent b)  Automatic Augmentation by using an API:
Augmentor \cite{bloice2017augmentor} by Bloice et al. is a software package which provides high level API in Python and Julia with a pipeline-based approach to perform various operations like rotation, translation, scaling, cropping, shearing etc. We used this package to augment our dataset to add following:

\begin{enumerate}
    \item Rotation
    \item Random Erasing (80 \% intact)
    \item Random cropping (80 \% intact)
    \item Flip (Left-Right \& Top-Bottom)
    \item Skew (Magnitude – 0.3)
\end{enumerate}

\begin{figure*}[htbp]
\centerline{\includegraphics[width=0.7\textwidth]{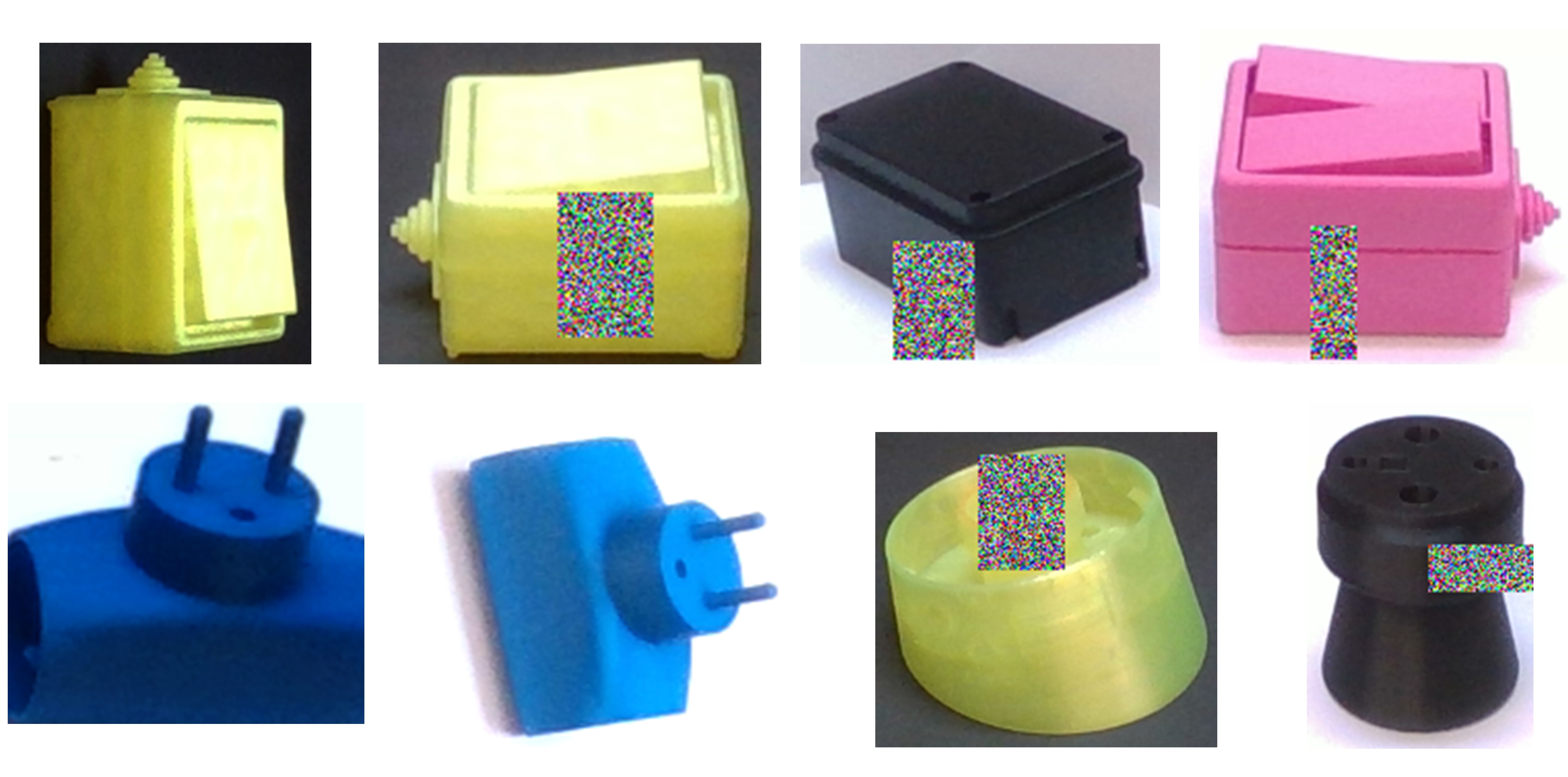}}
\caption{Results of Automatic Augmentation using Augmentor API}
\label{fig:Fig5}
\end{figure*}

A sample of the results of automatic augmentation are shown in ``Fig.~\ref{fig:Fig5}''.

Upon performing manual and automatic augmentation, we obtained a balanced dataset that contains similar number of images for all the classes. The new dataset comprised of 340,000 images belonging to all 30 classes. 

\medskip
\noindent \textbf{Part 2: Creation of 14 new datasets for feature comparison}
\medskip

From literature review, it could be inferred that for a textureless object, edges are the most important feature. In part 1, we augmented the data from 27,000 (Ground truth) to 340,000 (Augmented data). For obtaining the edge features, we implemented three different edge detectors on the ground truth data to obtain the corresponding edge features. 

\begin{enumerate}
   
    \item Canny Edge detector: By applying non-max suppression on the image gradient intensities, and applying double threshold, it finds the strong edges by suppressing the weak ones. 
    \item Holistically Nested edge detector (HED): Xie et al. \cite{xie2015holistically} developed a neural network, which uses a deep neural network to automatically learn hierarchical edge maps thereby identifying the edges of the objects in images.
    \item Prewitt operator: By convolving with a 3x3 kernel with the input image, it calculates the gradient of image intensity at each point by combining vertical and horizontal derivative approximations.
   
\end{enumerate}

Canny edge detector thin dotted lines along the edges of the object, but it was not continuous. HED provided edges of an object which gave thick outlines around the object. Prewitt gave thick outline for the object and thin curvy lines like canny for the inside part of the object. Thus, we created 3 datasets of \~340,000 images containing Canny features, HED features and Prewitt features. We also considered Sobel Edge Detector but since it gave a similar result as Canny, it was not implemented.

\begin{figure*}[htbp]
\centerline{\includegraphics[width=0.7\textwidth]{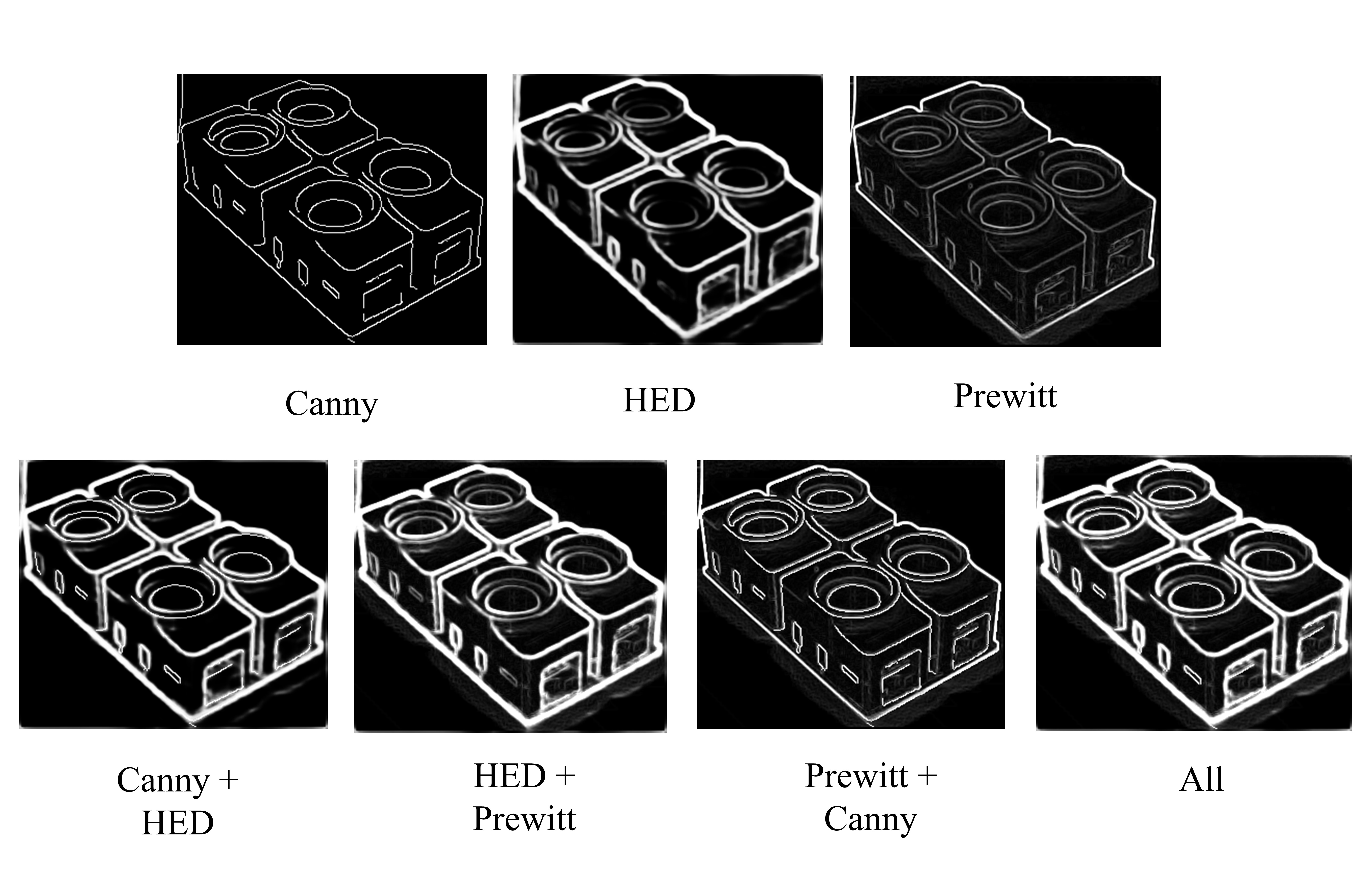}}
\caption{Sample of Features Only and Feature Combination dataset}
\label{fig:Fig6}
\end{figure*}

The next step was to combine the said features for all permutations to observe if they can perform better. First, we combined Canny (thin curvy outlines) with HED (thick outlines) which yielded us good results in the form of well-defined clear outlines of the edges. We then combined HED (thick outlines) with Prewitt (thin curvy outlines) which produced good results, but the outlines were not as well-defined as Canny-HED, especially the inner edges. For the third experiment, we implemented a combination of Canny (thin curvy lines) and Prewitt (thin curvy lines) which gave us thin outlines as well as inner lines of the object. Finally, the combination all three edge detectors – Canny, HED and Prewitt which yielded us thicker and more prominent edges for the object. So, we now had a total of 7 datasets, each containing ~340,000 images of features and feature combinations. A sample from these 7 datasets are shown in  ``Fig.~\ref{fig:Fig6}''.

\begin{figure*}[htbp]
\centerline{\includegraphics[width=0.8\textwidth]{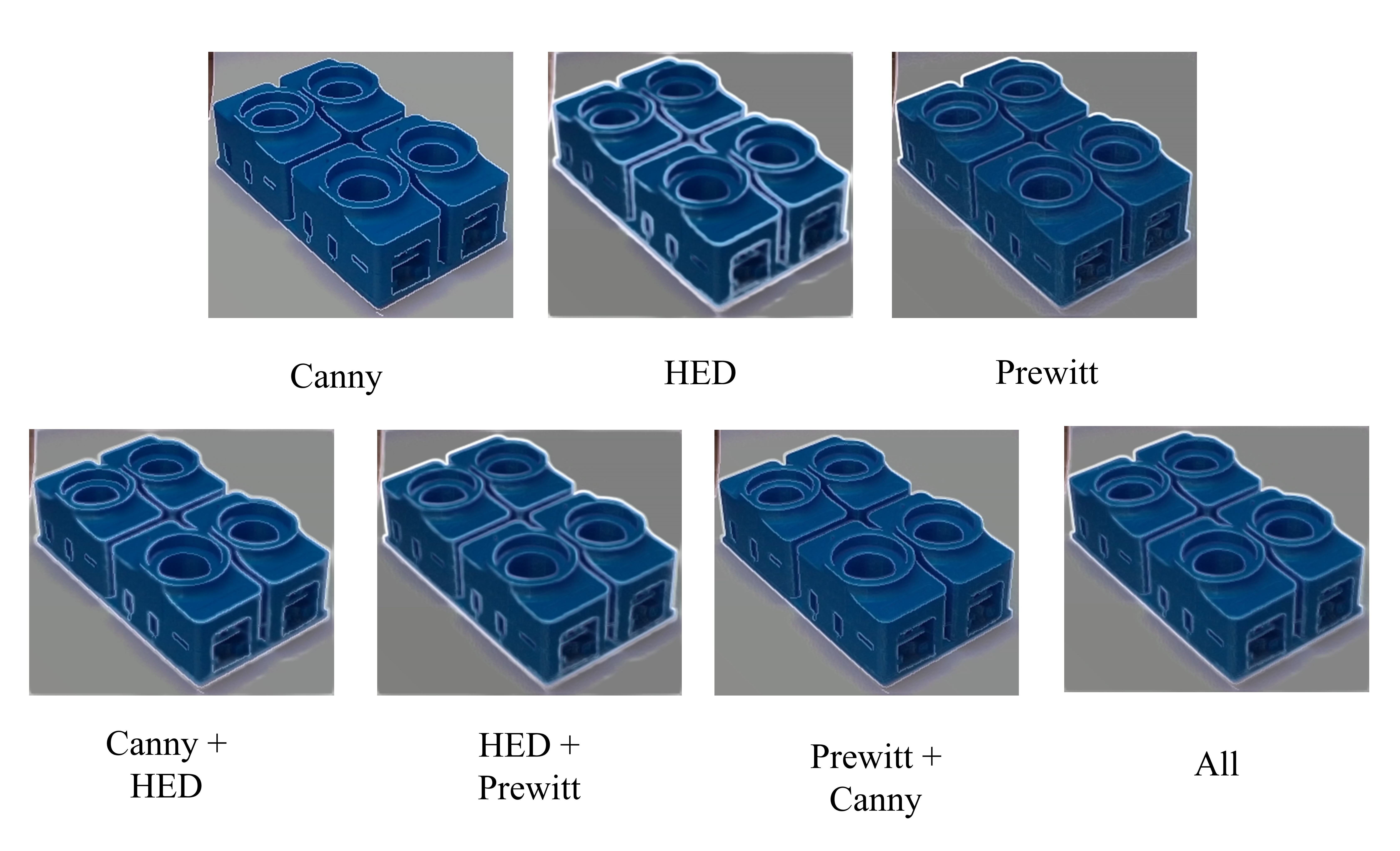}}
\caption{Sample of RGB images enhanced with Features Only and Feature Combination}
\label{fig:Fig7}
\end{figure*}

To retain the original RGB channel features, we performed overlay of above edges and edge combinations on the original RGB images to get edge enhanced RGB images as seen on  ``Fig.~\ref{fig:Fig7}''.

Hence, in total we created 15 datasets containing \~340,000 images each to amass a total of 5 million images along with bounding boxes and their annotations. 

\medskip
\noindent \textbf{Part 3: Training and Testing on Multiple Classifiers}
\medskip

Our research is focused on understanding the significance of edge features for textureless objects and to see if edges or edge enhanced pixel intensities can help improve recognition results. Hence, we planned to train our datasets on multiple models for comparing its performance and to have an unbiased result irrespective of choice of model.

\noindent \textit{1) Model Selection}
\medskip
Different models were considered based on the following criteria:
\begin{itemize}
    \item Support incremental learning as we have large dataset (15 GB per dataset)
    \item Faster training (\~45 minutes per dataset), as we had 15 datasets to train and predict
    \item Follows similar input/output dimensions for training
    \item Have similarity in input pre-processing
    \item Can be implemented efficiently inside a loop without duplicating lines of codes
\end{itemize}

With the above selection criteria in mind, we selected to train 4 models from sci-kit learn library:
\begin{enumerate}
    \item Stochastic Gradient Descent (SGD): It helps implement linear SVM or Logistic Regression based on the loss type selected, on a large scale. It performs really well especially when the data is sparse, which is the case of our dataset, as we are comparing edge features, which will eventually result in a nxn sparse matrix with only edge pixel locations having ‘1’ and all others ‘0’. We used this model with loss type ‘log’ for Logistic Regression.
    \item Perceptron: Perceptron is a simple classification algorithm which is faster than SGD, especially with ‘hinge’ loss and results in comparatively less sparser models. It does not require a learning rate and is not regularized; it learns from its own mistakes. Hence, it is easier to perform hyperparameter tuning with this model.
    \item Passive-Aggressive Classifier with hinge loss: Passive-Aggressive algorithms are used mainly for large scale learning. It does not need learning rate, but has regularization parameter. We have tried the classifier on both hinge loss and squared hinge loss as it would give us an understanding on what works better for our data
    \item Passive-Aggressive Classifier with squared-hinge loss
\end{enumerate}

\noindent \textit{2) Data Preprocessing}
\medskip

We have created 15 datasets: 7 datasets with edges and edge combinations, which are binary images and 7 datasets with edge-enhanced RGB images. Before sending the inputs to the model we have performed following preprocessing steps:

\begin{enumerate}
    \item Created class labels for each input image
    \item Created feature matrix for input images as a numpy array, with each row corresponding to each input image
    \begin{enumerate}
        \item For edges only inputs, we did not apply any initial image processing, but for edge enhanced RGB images, we initially converted them to grayscale
        \item The result is then resized to 200x200 to have match the dimensions for all input images
        \item The resized input is then flattened from 200X200 to 40000X1 array
    \end{enumerate}
    \item From step 1 and 2, we created each dataset with X as input feature matrix of size 340000 X 40000 x 1 and y as output labels with size 340000 X 1 x 1
\end{enumerate}

\medskip
\noindent \textit{3) Fitting and Training the model in batches}
\medskip

Once we have the input dataset, we keep 20\% data aside as unseen data for final testing. The remaining 80\% is then passed to the model Pipeline as batches which gets divided into 75\% and 25\% to be used for training and validation. 

\begin{figure*}
\centering
{\includegraphics[width=0.9\textwidth]{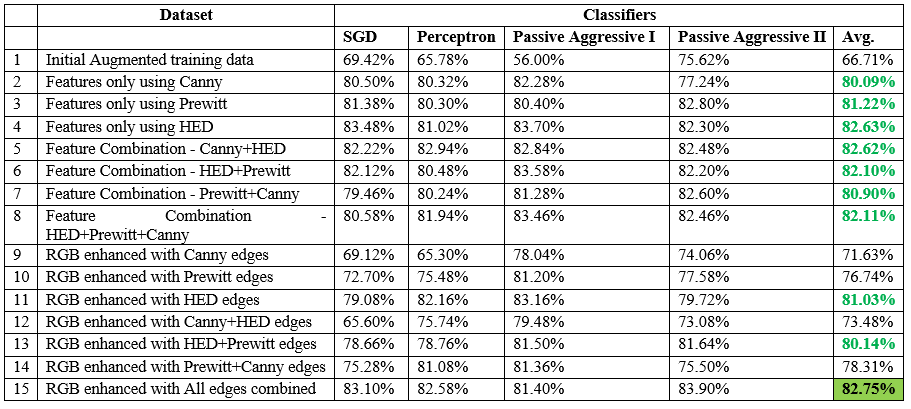}}
\caption{Comparison of Accuracy for 15 datasets across 4 classifiers}
\label{fig:Fig8}
\end{figure*}

\begin{figure*}[htbp]
\centerline{\includegraphics[width=0.9\textwidth]{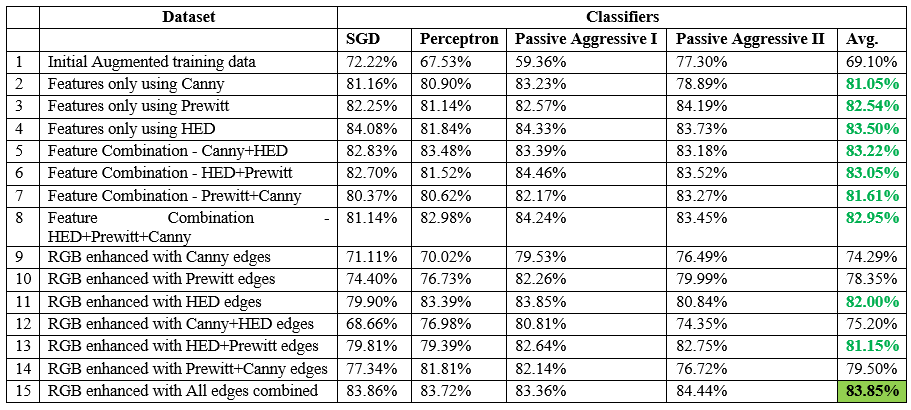}}
\caption{Comparison of F1 Score for 15 datasets across 4 classifiers}
\label{fig:Fig9}
\end{figure*}

\begin{figure*}[htbp]
\centerline{\includegraphics[width=0.8\textwidth]{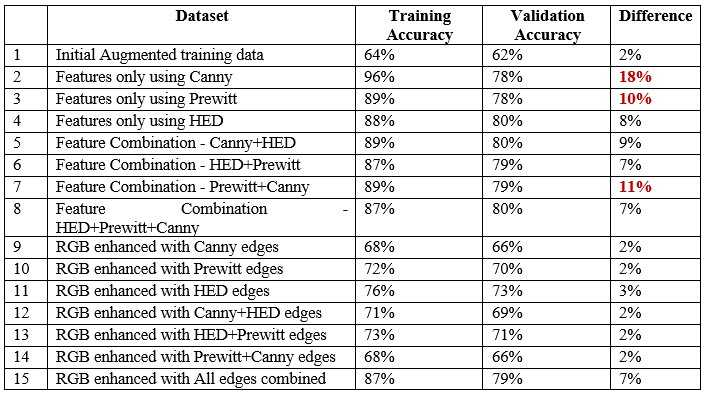}}
\caption{Comparison of Progressive Training and Validation accuracy to evaluate Overfitting for 15 datasets across 4 classifiers}
\label{fig:Fig10}
\end{figure*}

A batch of 5000 inputs are passed in every iteration into the Pipeline, which has a StandardScaler() and Model (). The StandardScaler() helps to standardize the input features and normalize the values  to be between 0 and 1. The standardized features are then passed to each model for training and validation. The progressive training and validation accuracies are noted and compared. The final model performance is noted and compared for Accuracy in  ``~\ref{fig:Fig8}'' and F1 score (balance between precision and recall) in ``~\ref{fig:Fig9}'' . To understand if the model is overfitted on any specific dataset, we have captured the difference between average of progressive testing and validation accuracies in ``~\ref{fig:Fig10}'' . 

\noindent \textit{Analysis and Result on Model Performance} 

\begin{enumerate}
    \item The best performance across all datasets for Accuracy as well as F1 score was presented by 15th  dataset which is the one with RGB images enhanced with all edges combined
    \item Features only and its combination dataset has shown considerably higher accuracy and F1 score for all classifiers compared to feature enhanced RGB versions. But it is to be noted that as per table 3, features only dataset has overfitted the model compared to its feature enhanced version, because of the high level of sparseness in the input feature matrix. 
    \item Across all the three edge detectors used for comparison, edges features using HED technique performed the best as a single feature and in combination whereas Canny scored the lowest scores 
\end{enumerate}

\medskip
\noindent \textit{4) Predicting on test dataset} 
\medskip

We have trained 4 models on 15 datasets, which is a total of 60 trained models. We test all of them on two different datasets: 1) Unseen images which are similar to the training dataset 2) Random images taken using a phone camera with different backgrounds.

\noindent \textit{a. Performance on Unseen Dataset with White background}

\begin{figure*}[htbp]
\centerline{\includegraphics[width=0.9\textwidth]{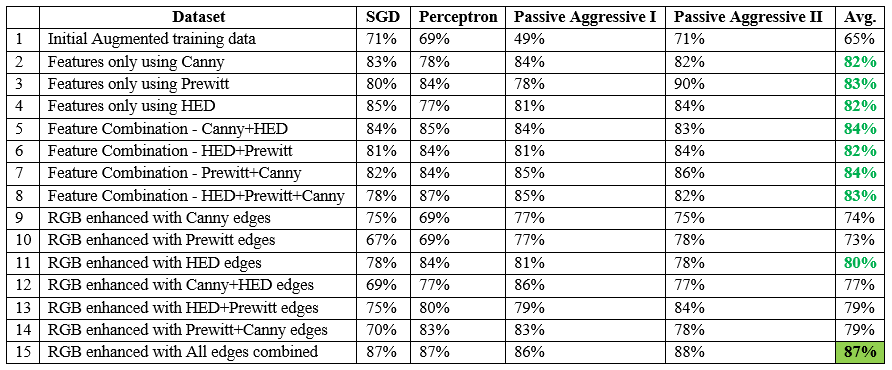}}
\caption{Comparison of model performance on test set 1 with white background}
\label{fig:Fig11}
\end{figure*}

As seen on ``~\ref{fig:Fig11}'' the model performance on the dataset with white background was similar to what was expected from the model performance during training.

\begin{enumerate}
    \item The best performance across all datasets for accuracy was presented by 15th  dataset which is the one with RGB images enhanced with all edges combined
    \item Features only dataset has shown considerably higher accuracy for all classifiers compared to Feature enhanced RGB versions.
    \item Unlike training model performance, all features performed almost equally well for features only dataset
\end{enumerate}

\noindent \textit{b. Performance on random dataset with different background}

As seen on ``~\ref{fig:Fig13}'' the model performance on the random dataset with different background was different from test set 1. 

\begin{figure}[htbp]
\centerline{\includegraphics[width=0.45\textwidth]{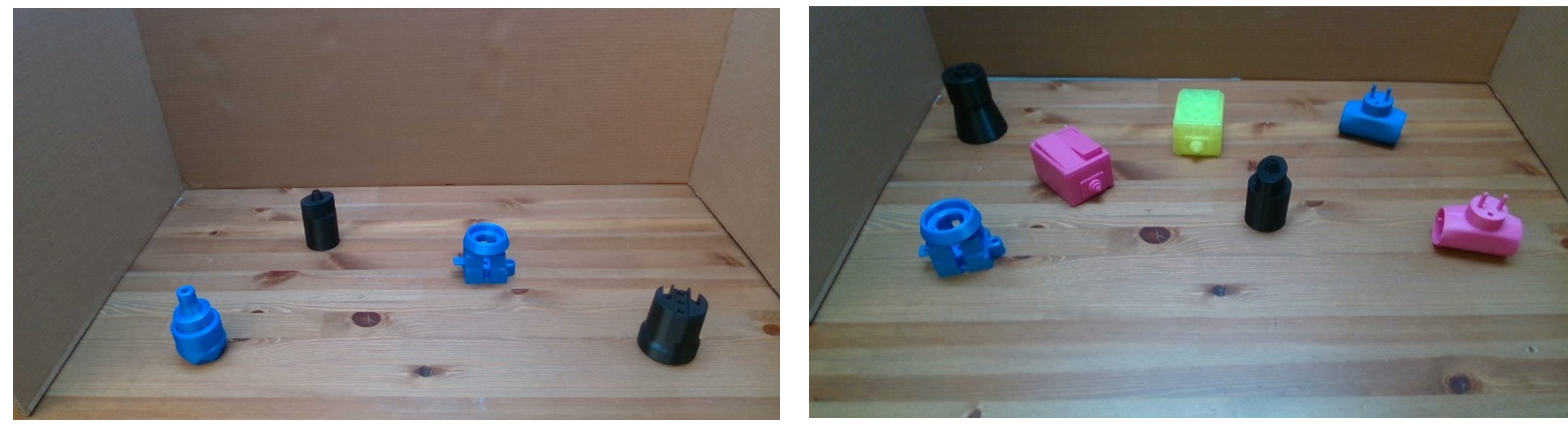}}
\caption{Sample images from test set 2}
\label{fig:Fig12}
\end{figure}

\begin{enumerate}
    \item The best performance across all datasets for accuracy was presented by 15th  dataset which is the one with RGB images enhanced with all edges combined.
    \item Even though there was an improvement in accuracy from initial augmented data to feature enhanced images, the overall accuracy was way lower for test set 2 compared to test set 1. The reason for this could be that since the model is trained on the images with white background, it may not be performing well for the ones with different background. The model only understands the pixels where there are edges. For a test image with background, its edges will also appear in the feature matrix, which makes it difficult for model to understand and predict. 
    \item HED features performed the best as a single feature and in feature combinations. 

\end{enumerate}

\begin{figure*}[htbp]
\centerline{\includegraphics[width=0.9\textwidth]{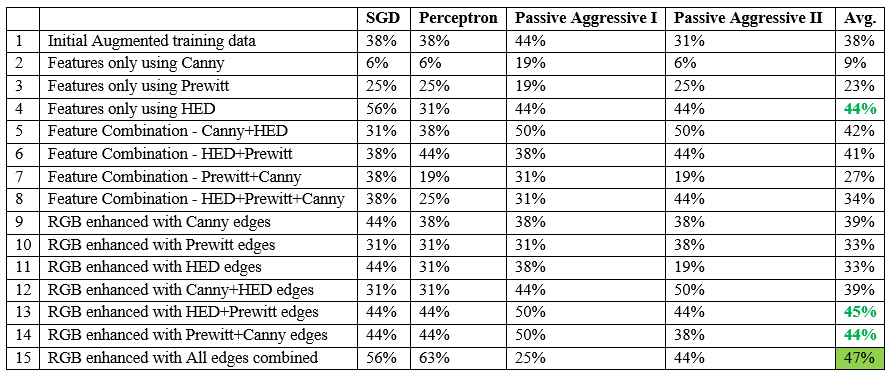}}
\caption{Comparison of model performance on test set 2 with different background}
\label{fig:Fig13}
\end{figure*}

\section{Summary of Results}

We trained and tested 4 models on 15 datasets with features only, features combination and RGB images enhanced with the same. The results show that the original image enhanced with combination of all three edges performed the best overall. Comparing the three edge detectors, performance on datasets with HED edges as single feature and its combination performed better. This could be attributed to the clear, thick edge features given by HED.  

\section{Conclusion and Future Work} 

The project aimed at creating a robust, balanced augmented dataset from initial set of 27,000 ground truth data of textureless objects. Using that as baseline, we created 14 other datasets each of ~340,000 images by using 3 edge detectors, Canny, Prewitt and HED. The initial seven were 3 single features and 4 of its combination dataset. The next seven were RGB images enhanced with the initial seven dataset. The experiments proved that the best performance was presented by the RGB images enhanced with all features combined. HED features proved to be performing the best among all features. Since the initial ground truth was with white background, the models performed really well on images with white background, whereas it could not perform when there was a different background. In future work, we would like to perform the experiments that can generalize the comparison between features irrespective of background or clutter. This could be achieved by adding background and clutter to the training set or by removing background or clutter from the test set. 

\section{Acknowledgement}

The authors would like to thank our mentor, Gabriel Lugo Bustillo, Phd candidate, dept. of Computing Science, University of Alberta for his guidance and feedback throughout the research. We would also like to thank our advisors Dr. Irene Cheng and Dr. Anup Basu for their support and motivation to bring out the best in our research. We would also like to thank the researchers of previous work in the field of Textureless Object Recognition, which became the starting point for our research.

\section{Resources}

Source Code:  \href{https://tinyurl.com/v7bbrkb}{https://tinyurl.com/v7bbrkb} 

\section {Author Biography}
\begin{IEEEbiography}[{\includegraphics[width=1in,height=1.25in,clip,keepaspectratio]{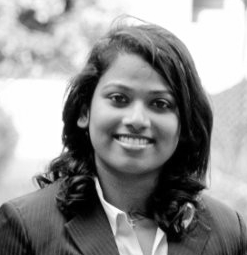}}]%
{Frincy Clement}
is an Artificial Intelligence enthusiast who is currently pursuing her MSc. in Computing Science at the University of Alberta. She received her BTech in Computer Science and Master in Business Administration from top universities in India. She has 6+years of work experience and is a seasoned software developer with experience working in several application as well as research-based projects in AI.
\end{IEEEbiography}

\begin{IEEEbiography}
[{\includegraphics[width=1in,height=1.25in,clip,keepaspectratio]{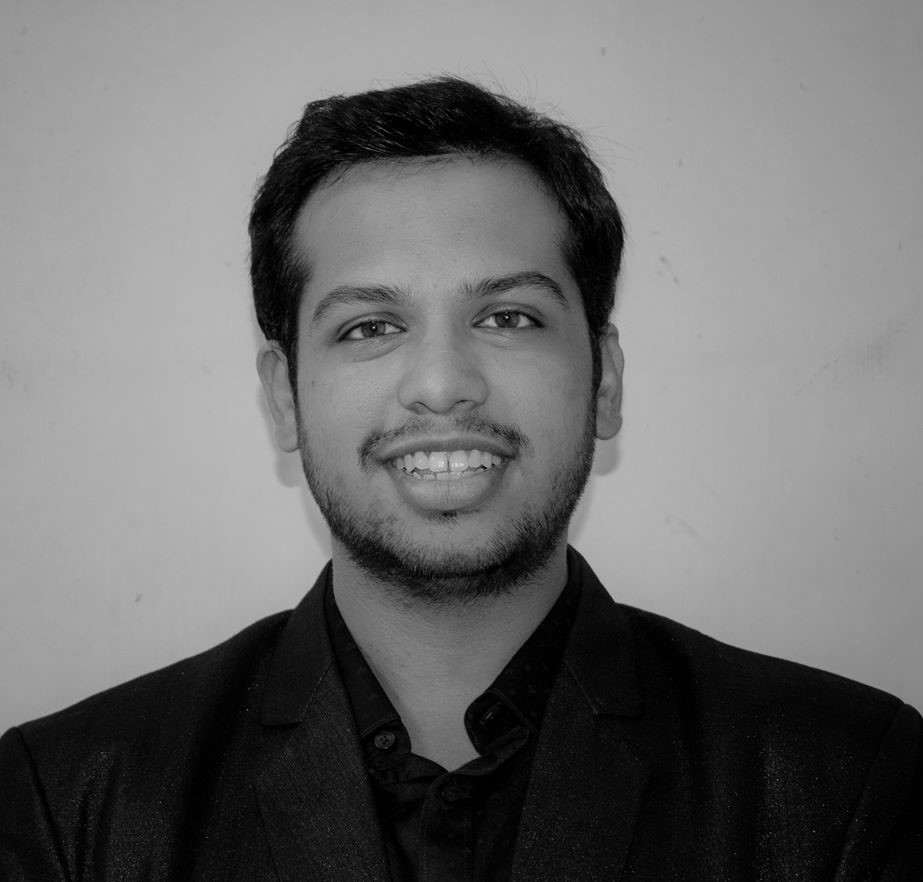}}]%
{Kirtan Shah}
is a student pursuing Masters in the field of Computing Science at University of Alberta. He completed his Bachelor’s in the field of Computer Engineering in India. He has worked on projects that integrate IOT application for the real-world needs during his bachelor’s. His interest is to apply various techniques to solve real world problems. He likes to play chess as well.
\end{IEEEbiography}

\begin{IEEEbiography}
[{\includegraphics[width=1in,height=1.25in,clip,keepaspectratio]{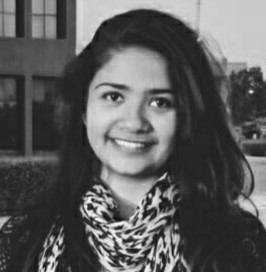}}]%
{Dhara Pancholi}
is currently a MSc. in Computing Science specializing in Multimedia at University of Alberta. She completed her bachelor's degree in Computer Science from India. Apart from different subjects that she learned, her interest lies in computer graphics and animation. She has also worked on different IOT projects in her bachelors and currently pursuing 3 projects at university of Alberta as course projects.
\end{IEEEbiography}

\bibliographystyle{ieeetran}
\bibliography{References}

\section{Appendix}

\begin{figure*}[htbp]
\centerline{\includegraphics[width=0.6\textwidth]{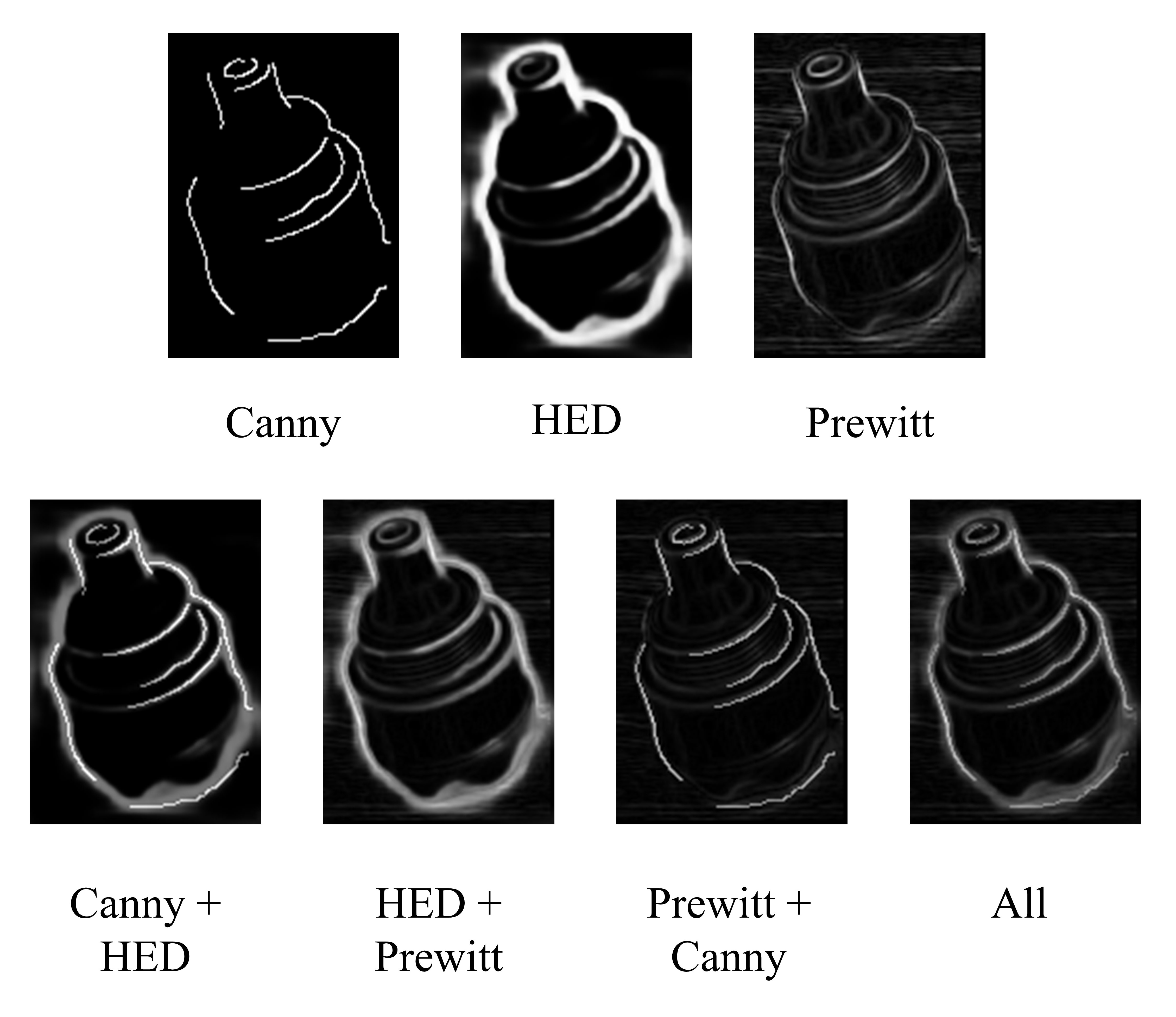}}
\caption{Sample of test set 2 after initial preprocessing for features/feature combinations}
\label{fig:Fig14}
\end{figure*}

\begin{figure*}[htbp]
\centerline{\includegraphics[width=0.6\textwidth]{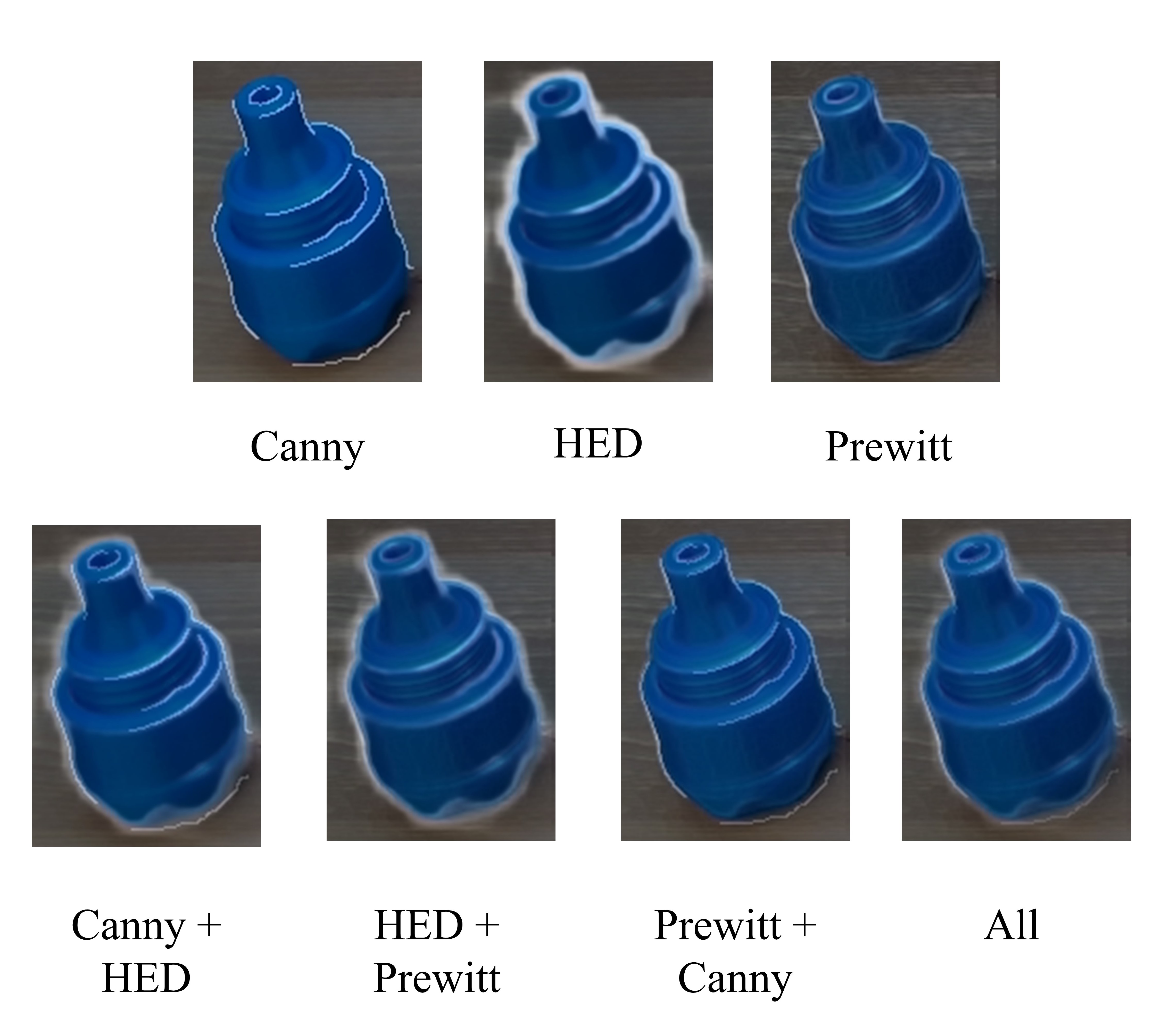}}
\caption{Sample of test set 2 after initial preprocessing for RGB enhanced with features/feature combinations}
\label{fig:Fig15}
\end{figure*}

\end{document}